\documentclass{article}

\PassOptionsToPackage{numbers, compress}{natbib}

\usepackage[dblblindworkshop, final]{neurips_2026}
\workshoptitle{Self-Evolving Diversity-Driven Search for Robust AI Systems}

\usepackage[utf8]{inputenc} 
\usepackage[T1]{fontenc}    
\usepackage[hidelinks]{hyperref}       
\usepackage{url}            
\usepackage{booktabs}       
\usepackage{amsfonts}       
\usepackage{nicefrac}       
\usepackage{microtype}      
\usepackage{xcolor}         
\usepackage{graphicx}
\usepackage{algorithm}
\usepackage{algpseudocode}
\usepackage{caption}
\usepackage{amsmath}
\usepackage{caption}
\usepackage{array}

\title{Design and Tuning of Multi-Agent Systems Using Bayesian Evolutionary Optimization}

\author{
  Dominik Jehle \\
  \texttt{jehled@cs.uni-freiburg.de}\\
  University of Freiburg \\
  Freiburg im Breisgau, Germany
  \And
  Tom Zehle \\
  \texttt{tom.zehle@tue.ellis.eu}\\
  University of Freiburg \\
  Freiburg im Breisgau, Germany
}

\begin{document}

\maketitle

\begin{abstract}
Designing effective multi-agent systems (MAS) remains a challenging and resource-intensive task. We propose a Bayesian Evolutionary Optimization framework for automatically discovering task-specific MAS topologies. The approach combines evolutionary operators to generate diverse workflow graphs with Bayesian optimization to focus cost intensive evaluations on promising candidates. In contrast to LLM-based candidate selection, our approach uses a graph-aware Gaussian Process and Expected Improvement to provide an explicit selection criterion during evolutionary search. We evaluate the method on BigCodeBench for code generation and on a tabular preprocessing benchmark using TabPFN. On BigCodeBench, evolved architectures achieve the highest overall performance among the evaluated agent systems. On temporal tabular datasets, optimized MAS preprocessing improves over the default TabPFN model on several tasks, while remaining competitive on IID datasets. Overall, the results demonstrate the potential of Bayesian selection for guiding evolutionary MAS search under expensive evaluation budgets.

\end{abstract}

\section{Introduction}

Multi-agent systems (MAS) have demonstrated strong capabilities across a wide range of complex tasks, including coding and data analysis \cite{fourney2024magenticonegeneralistmultiagentsolving, tran2025multiagentcollaborationmechanismssurvey}. By combining multiple specialized agents, MAS can leverage complementary expertise to solve problems that are often beyond the capabilities of a single agent. However, different tasks may benefit from different agent roles and workflow structures, making it challenging and resource-intensive to manually design an effective, task-specific MAS topology \cite{li2024survey, he2025llm}. Recent research has shown the potential of combining large language models (LLMs) with evolutionary algorithms. In these approaches, LLMs are used to generate and refine candidate solutions, while evolutionary algorithms provide an efficient mechanism for exploring the search space and discovering increasingly effective system designs \cite{Yu2024LLMEA}.

To address the high cost of manually designing task-specific multi-agent systems, we propose a \textbf{Bayesian Evolutionary Optimization} method for automated MAS design. Our approach combines evolutionary algorithms to generate and diversify candidate MAS topologies with Bayesian optimization to efficiently identify the most promising architectures for evaluation.

\paragraph{Contributions.} I) We propose the \textbf{Bayesian Evolutionary Optimization Pipeline}, a comprehensive MAS optimization framework that combines evolutionary operators with a specialized Bayesian optimization strategy for candidate selection. The pipeline is based on the Promptolution framework \cite{zehle2026promptolution} and is evaluated on two distinct benchmark settings. II) We introduce a \textbf{tabular data benchmark} that evaluates MAS-based preprocessing for tabular datasets. The benchmark assesses whether MAS optimization can improve data preparation for state-of-the-art tabular prediction models such as TabPFN \cite{hollmann2023tabpfn, grinsztajn2026tabpfn3technicalreport}.

\section{Related Work}

The automated design and optimization of multi-agent systems (MAS) has recently gained significant research interest. ADAS \cite{hu2025automated} uses a meta-agent to iteratively improve existing systems and generate new MAS architectures. By building on previously discovered high-performing systems, ADAS outperforms conventional prompting and manually designed agent systems \cite{hu2025automated}. MASS \cite{zhou2026multiagentdesignoptimizingagents} similarly optimizes MAS iteratively by separately optimizing graph topologies and agent prompts before combining them. It achieves strong performance compared to other automated frameworks such as AFLOW \cite{zhang2025aflow}.

Evolutionary approaches have also been applied to MAS optimization. EVOAGENT \cite{yuan2025evoagent} evolves MAS topologies through mutation and crossover, with an LLM guiding candidate selection, and outperforms prompting strategies and agent frameworks such as AgentVerse and AutoAgent. EVOMAS \cite{hu2026evomas} combines evolutionary operators with LLM-based selection and achieves superior performance over single-agent systems, handcrafted architectures, and other automated methods across benchmarks including SWE-bench and WorkBench. Beyond LLM-based MAS optimization, GAPPO \cite{fang2025efficient} integrates genetic algorithms with reinforcement learning through mutation, crossover, and selection.

While existing evolutionary MAS approaches primarily rely on LLM-based selection, we investigate Bayesian optimization as an alternative selection mechanism. Rather than asking an LLM to judge candidate architectures, our method explicitly models previously evaluated graph topologies and uses Expected Improvement to select candidates for evaluation. This provides an explicit, model-based selection criterion, reducing reliance on
stochastic LLM-based judgments.

\section{Method}
\begin{figure}[h]
\centering
\includegraphics[width=0.9\textwidth]{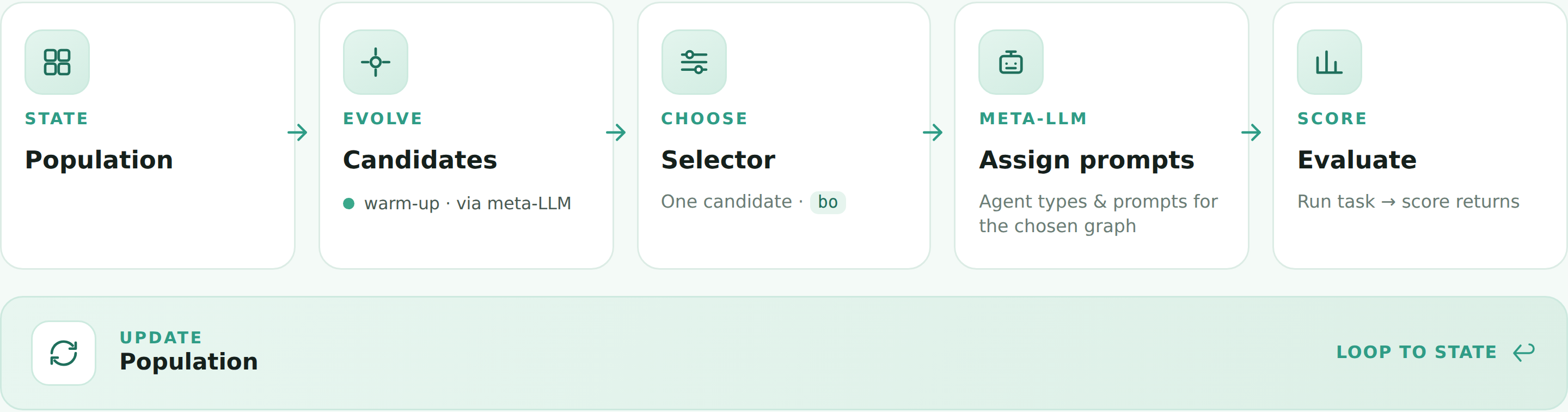}
\caption{Overview of the proposed optimization pipeline: Candidate MAS topologies are generated through evolutionary expansion, selected via Bayesian optimization, instantiated with agent types and prompts, and evaluated iteratively.}
\label{fig:evolutionary_pipeline}
\end{figure}

\begin{algorithm}
\caption{Bayesian Evolutionary Optimization for MAS Design}
\label{alg:mas_optimization}
\begin{algorithmic}[1]
\Require Target task $T$, initial population size $P$, candidates per iteration $N$
\Ensure Best-performing MAS architecture

\State Generate an initial population of workflow graphs using an LLM
\State Evaluate the initial population
\State Add evaluated graphs and their scores to the population

\While{optimization budget is not exhausted}
    \State Generate $N$ candidate graphs using mutation and crossover
    \State Compute graph similarities to previously evaluated graphs
    \State Fit a Gaussian Process using the observed evaluation scores
    \State Compute Expected Improvement (EI) for each candidate
    \State Select candidate $x^*$ with the highest EI
    \State Assign agent types and prompts to $x^*$
    \State Evaluate the resulting MAS on task $T$
    \State Add $x^*$ and its evaluation score to the population
\EndWhile

\State \Return Best-performing evaluated MAS architecture
\end{algorithmic}
\end{algorithm}
The proposed method extends the Promptolution framework \cite{zehle2026promptolution} with a complete optimization pipeline for multi-agent systems (MAS). The pipeline combines evolutionary exploration with Bayesian selection to efficiently discover task-specific MAS architectures. Evolutionary operators generate diverse candidate structures, while Bayesian optimization provides an explicit selection criterion for deciding which candidates should undergo expensive end-to-end evaluation. This separation allows the search to maintain structural diversity without requiring every generated topology to be evaluated. The complete procedure is summarized as Pseudocode in Algorithm~\ref{alg:mas_optimization}. 

\paragraph{Warm-Up Phase and Evolutionary Expansion.} 
During the warm-up phase, an initial population of workflow graphs is generated using an LLM as the basis for evolutionary optimization. The number of nodes is randomly determined for each graph, producing workflows of varying sizes and topologies.
The evolutionary expansion phase is driven by the application of mutation and crossover operators.
\begin{figure}[h]
\centering
\begin{minipage}{0.38\textwidth}
\centering
\includegraphics[width=\linewidth]{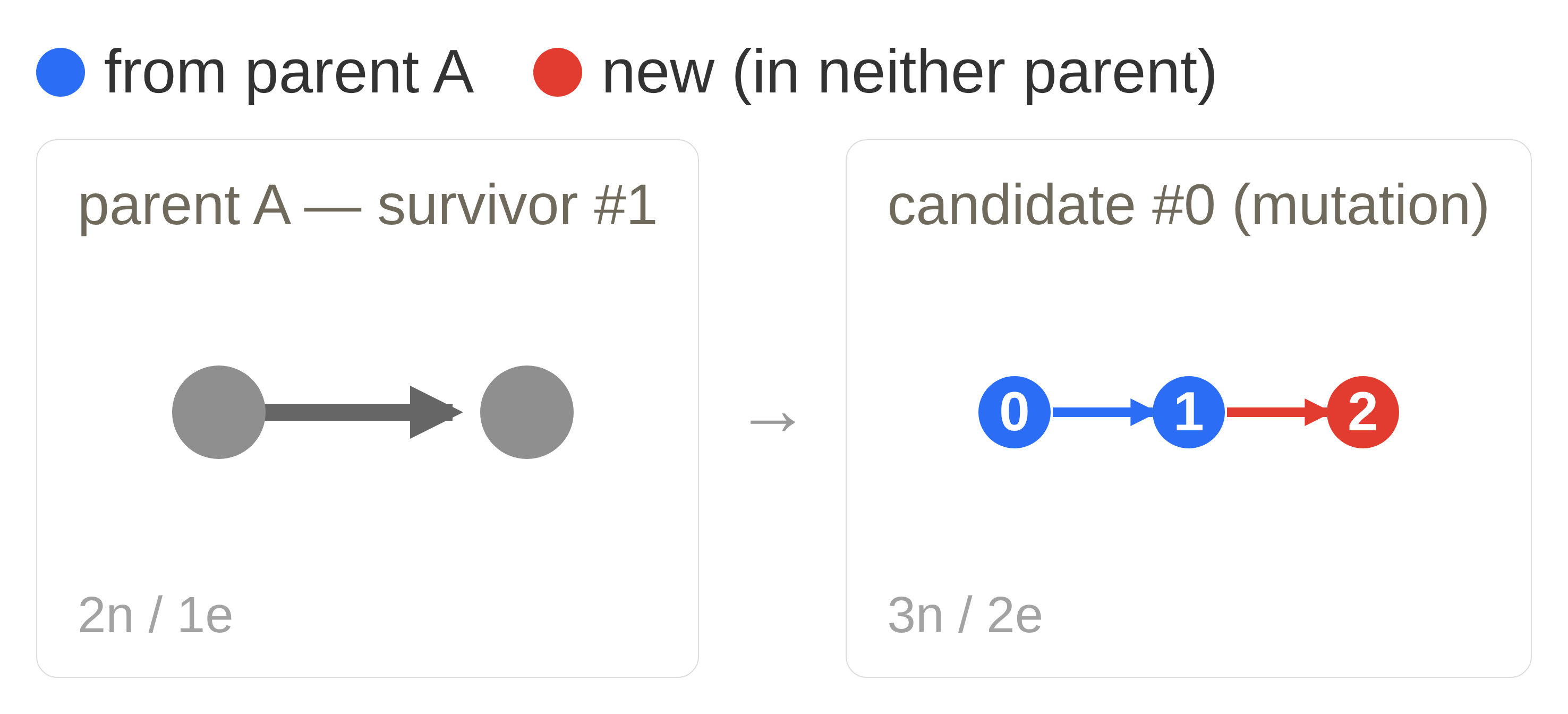}
\caption{Example of a mutation using the \emph{add node} operator. The other mutation operators are \emph{add edge}, \emph{remove node}, \emph{remove edge}, and \emph{rewire edge}.}
\label{fig:mutation}
\end{minipage}
\hfill
\begin{minipage}{0.54\textwidth}
\centering
\includegraphics[width=\linewidth]{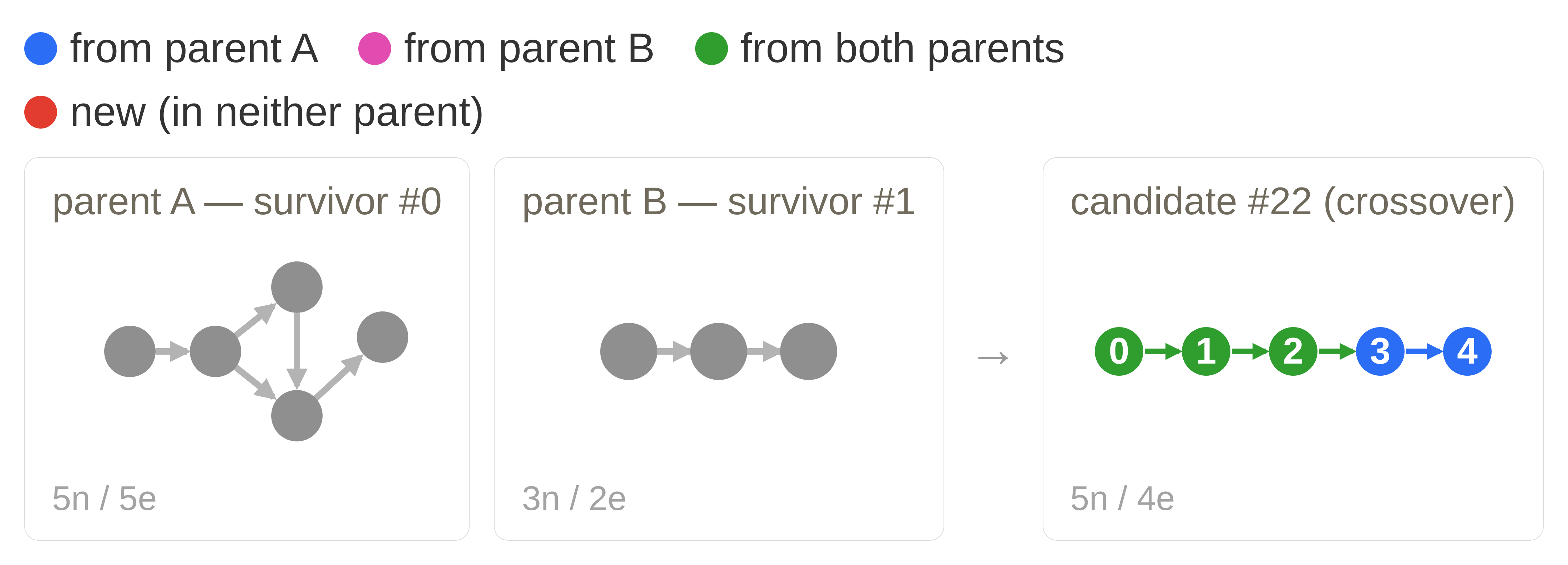}
\caption{Example of a crossover operation. Nodes and edges are sampled from both parent graphs to construct the offspring. }
\label{fig:crossover}
\end{minipage}
\end{figure}
For mutation, one of five mutation operators is selected at random and applied to a randomly chosen graph from the current population. This produces a new candidate graph, as illustrated in Figure~\ref{fig:mutation}.
For crossover, two parent graphs are selected from the population. The size of the offspring graph is randomly chosen between the sizes of the two parent graphs. Nodes and edges are then sampled from both parents with a predefined probability to construct the offspring, as illustrated in Figure~\ref{fig:crossover}.

\paragraph{Bayesian Selection Process}
After generating $N$ candidate graph topologies, only a single candidate is selected for evaluation due to the high computational cost of the evaluation process. To identify the most promising candidate, we employ Bayesian optimization with Expected Improvement (EI) \cite{frazier2018tutorial, frazier2018bayesian, shahriari2015taking}. Since the optimization operates on graph structures, graph similarity is measured using a normalized, type-aware graph edit distance (GED) \cite{gao2010survey}. Let $V_x$ and $E_x$ denote the vertex and edge sets of graph $x$. The similarity kernel is defined as:
\begin{equation}
k(x, x') = 1 - \frac{\mathrm{GED}(x, x')}
{|V_x| + |E_x| + |V_{x'}| + |E_{x'}|}.
\end{equation}

Using this kernel, a Gaussian Process estimates the expected performance $\mu(x)$ and predictive uncertainty $\sigma(x)$ of each candidate based on previously evaluated graph topologies. Expected Improvement is designed to balance exploration and exploitation by estimating the potential improvement over the best observed performance $f_{\text{best}}$. It is defined as:
\begin{equation}
\mathrm{EI}(x) =
(\mu(x)-f_{\mathrm{best}}-\xi)\Phi(z)
+\sigma(x)\phi(z),
\quad
z=\frac{\mu(x)-f_{\mathrm{best}}-\xi}{\sigma(x)}
\end{equation}

Here, $\Phi$ and $\phi$  denote the standard normal CDF and PDF, respectively, while $\xi$ controls exploration. The candidate with the highest Expected Improvement is selected for evaluation (details in Appendix~\ref{Bayesian Selection Process}).
\begin{equation}
x^{\star} = \operatorname*{arg\,max}_{x \in C} \mathrm{EI}(x).
\end{equation}

\paragraph{Prompt Assignment}
After selecting a graph topology, each node is instantiated as an agent by assigning an agent type and system prompt. The framework initially provides three base types: \emph{Planner}, \emph{Executor}, and \emph{Validator}. An LLM first analyzes the task and generates additional task-specific agent types. Then, a three-stage pipeline configures each node: a scheduler LLM assigns agent types based on the graph and task, another LLM generates corresponding system prompts, and the final configuration combines agent type, prompt, and required tools. The resulting graph is a fully instantiated multi-agent system ready for evaluation.

\section{Experimental Setup}
We evaluated our method on two benchmarks. First, we developed a custom tabular benchmark to assess whether MAS optimization can improve preprocessing for downstream prediction. The objective is to optimize the preprocessing pipeline such that the resulting data maximizes the predictive information available to the model. Second, we evaluated our approach on BigCodeBench \cite{zhuo2025bigcodebenchbenchmarkingcodegeneration} as a traditional MAS benchmark. As baselines, we considered a Single Agent system and a manually designed \emph{Planner} $\rightarrow$ \emph{Executor} $\rightarrow$ \emph{Validator} MAS.

Across all experiments, we used Gemma 4 31B \cite{gemma4_modelcard} as both the meta-model and the underlying language model for all agents. To improve reproducibility, we fixed the sampling temperature to 0.25 \cite{LI2025242, grullon2026llm}. All benchmarks used separate training and test sets.

\section{Results}

\begin{figure}[h]
\centering
\includegraphics[width=0.9\textwidth]{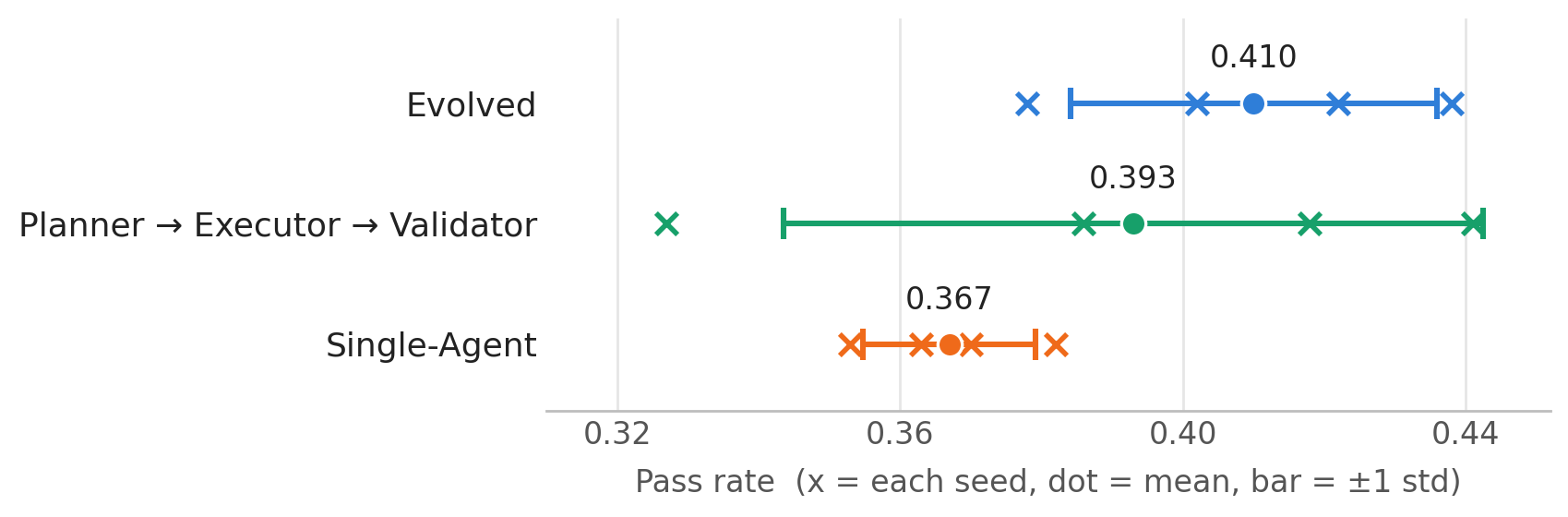}
\caption{Performance on BigCodeBench aggregated over all four evaluation seeds. The figure shows the results for all four seeds together with the mean performance and its standard deviation.}
\label{fig:tab_graph}
\end{figure}

\paragraph{BigCodeBench}
For the evaluation on BigCodeBench \cite{zhuo2025bigcodebenchbenchmarkingcodegeneration}, we used four different random seeds, each containing 300 benchmark tasks. Figure~\ref{fig:tab_graph} shows that the evolved architecture achieved the highest mean performance across the seeds (0.410), compared with the Planner–Executor–Validator architecture (0.393) and the Single Agent baseline (0.367). This corresponds to an absolute improvement of 0.017 over the manually designed multi-agent system and 0.043 over the Single Agent baseline. The evolved architecture improved mean performance by 4.3\% over the Planner–Executor–Validator architecture and by 11.7\% over the Single Agent baseline across the four seeds. The results suggest that architecture evolution can identify effective agent configurations that improve code-generation performance beyond both manually designed multi-agent coordination and single-agent execution.

Table~\ref{tab:solved_overlap} complements these results by reporting the total number of unique tasks solved across all four seeds. Since tasks may overlap between seeds, each task is counted only once, resulting in 775 unique tasks. The evolved architecture solved the most unique tasks (295; 38.1\%), followed by the Planner–Executor–Validator architecture (290; 37.4\%) and the Single Agent baseline (265; 34.2\%). These results provide complementary evidence that the evolved architecture achieves both stronger average performance across seeds and broader task coverage.

\begin{table}[h]
\centering
\caption{Comparison of multi-agent architectures on BigCodeBench across 775 unique tasks.}
\label{tab:solved_overlap}
\begin{tabular}{@{}lrr@{}}
\hline
\toprule
\textbf{Architecture} & \textbf{Tasks Solved} & \textbf{Percentage} \\
\midrule
Evolved & 295 & 38.1\% \\
Planner-Executor-Validator & 290 & 37.4\% \\
Single Agent & 265 & 34.2\% \\
\bottomrule
\end{tabular}
\end{table}
\newpage
\paragraph{Tabular Prediction}
For the tabular data benchmark, we used TabPFN as the tabular predictor and included an additional baseline by evaluating the raw performance of TabPFN \cite{hollmann2023tabpfn, grinsztajn2026tabpfn3technicalreport}. We evaluated all methods in two scenarios using datasets from BeyondArena \cite{purucker2026iidgeneraltabularfoundation}. The first scenario consists of 20 IID regression datasets, while the second uses 5 temporal regression datasets. Both benchmark sets were selected from the 142 BeyondArena datasets, restricting the datasets to a maximum of 50,000 rows. All datasets are listed in Appendix~\ref{Datasets Tabular Prediction}. The results are shown in Table~\ref{fig:IDD_mean} and~\ref{fig:temporal_mean}.

\begin{table}[h]
\centering
\begin{minipage}{0.45\textwidth}
\centering
\captionof{table}{Mean \(R^2\) of the evolved architecture and baselines on 20 IID regression datasets.}
\begin{tabular}{@{}lr@{}}
\toprule
\textbf{Model} & \textbf{Mean $R^2$} \\
\midrule
\textbf{Evolved architecture} & \textbf{0.611} \\
Single Agent (no type) & 0.588 \\
(Planner $\rightarrow$ Executor $\rightarrow$ Validator) & 0.527 \\
\bottomrule
\end{tabular}
\label{fig:IDD_mean}
\end{minipage}
\hfill
\begin{minipage}{0.48\textwidth}
\captionof{table}{Mean \(R^2\) of the evolved architecture and baselines on 5 temporal regression datasets.}
\centering
\begin{tabular}{@{}lr@{}}
\toprule
\textbf{Model} & \textbf{Mean $R^2$} \\
\midrule
\textbf{Evolved architecture} & \textbf{0.544} \\
Single Agent (no type) & 0.473 \\
(Planner $\rightarrow$ Executor $\rightarrow$ Validator) & 0.351 \\
\bottomrule
\end{tabular}
\label{fig:temporal_mean}
\end{minipage}
\end{table}
On both IID and temporal regression datasets, the evolved architecture achieved the best performance among the evaluated agent-based systems, including the Single Agent baseline and the manually designed Planner–Executor–Validator architecture. It achieved a mean $R^2$ of 0.611 on IID datasets and 0.544 on temporal datasets, improving over the Single Agent baseline by 0.023 and 0.071 and over the Planner–Executor–Validator architecture by 0.084 and 0.193. These results highlight the potential of architecture evolution to identify more effective agent configurations for coordinating preprocessing and prediction, particularly on temporal regression tasks.

While raw TabPFN remains a strong baseline, the evolved architecture demonstrates potential for task-specific improvements through its learned preprocessing workflows. For example, it achieved higher predictive performance than raw TabPFN on two of the five temporal datasets. In terms of aggregate performance, the evolved architecture achieved a comparable mean $R^2$ on temporal datasets (0.544 vs. 0.535), although raw TabPFN performed slightly better on IID datasets (0.620 vs. 0.611). These results suggest that evolved preprocessing can be beneficial for certain tasks, while also highlighting opportunities to improve its robustness. Potential sources of performance degradation include hallucinated information, information loss caused by inappropriate feature transformations or type conversions, and unnecessary mathematical operations that introduce misleading information. Prompt optimization is a promising direction for improving the robustness of the preprocessing pipeline. The complete results are shown in Appendix~\ref{Additional Results Tabular Prediction}.

\section{Conclusion}
In this project, we proposed a novel approach for automatically discovering task-specific multi-agent system (MAS) topologies by combining evolutionary algorithms with Bayesian optimization. The evolved architectures achieve higher average performance than both a Single Agent baseline and a manually designed multi-agent system architecture on BigCodeBench and showed competitive performance on the tabular data benchmark.

\paragraph{Limitations and Future Work}
A key limitation is that topology optimization and prompt generation are optimized separately. While Bayesian optimization guides topology selection, prompt assignment may still affect final performance. Moreover, the GED-based similarity is not guaranteed to be positive semi-definite,
which may affect the behavior of the Bayesian selection (see Appendix~\ref{Bayesian Selection Process}). Future work could jointly optimize topologies and prompts, for example by incorporating prompt optimization methods such as Cantante \cite{zehle2026cantante}. Further evaluation across a broader range of tasks and domains is needed to assess the generalizability of the approach.

\section*{Acknowledgements}
Tom Zehle received funding by the European Union. This work was supported by the European Union's Horizon Europe research and innovation program under grant agreement No. 101214398 (ELLIOT).
Finally, we thank the reviewers for their constructive feedback and contribution to improving the paper.

\bibliographystyle{unsrtnat}
\bibliography{refs}


\newpage
\appendix

\section{Bayesian Selection Process}
\label{Bayesian Selection Process}
After generating $N$ candidate graph topologies during the evolutionary expansion phase, only a single candidate is selected for evaluation, as the evaluation process is computationally expensive.

To efficiently identify the most promising candidate, we employ Bayesian optimization with the Expected Improvement (EI) acquisition function \cite{frazier2018tutorial, frazier2018bayesian, shahriari2015taking}. Since the optimization operates on graph structures rather than continuous vectors, graph similarity is measured using a normalized, type-aware graph edit distance (GED) \cite{gao2010survey}. Let $V_x$ and $E_x$ denote the vertex and edge sets of graph $x$. The similarity kernel is defined as

\begin{equation}
k(x, x') = 1 - \frac{\mathrm{GED}(x, x')}
{|V_x| + |E_x| + |V_{x'}| + |E_{x'}|}
\end{equation}

Using this kernel, Gaussian Process regression estimates the posterior mean and variance for each candidate graph:

\begin{equation}
\mu(x_*)      = \bar{y} + k_*^{\top} K_y^{-1} (y - \bar{y})\
\end{equation}
\begin{equation}
\sigma^2(x_*) = k(x_*, x_*) - k_*^{\top} K_y^{-1} k_*
\end{equation}

where $K_y$ denotes the kernel matrix computed from all previously evaluated graphs, augmented with a small diagonal jitter term to ensure numerical stability. Furthermore, $k_* = [k(x_*, x_i)]_i$ represents the similarity between the candidate graph and all previously evaluated graphs, while $\bar{y}$ denotes the mean observed evaluation score.

The GED-based kernel is not positive semi-definite in general. In our
implementation, negative posterior variances were clamped to $10^{-12}$ rather than
corrected, which biases selection towards candidates dissimilar to evaluated graphs
and deviates from standard EI. A projection of the joint kernel matrix onto the PSD
cone would resolve this and is left for future work.

The Expected Improvement acquisition function evaluates each candidate relative to the best-performing graph observed so far, $f_{\text{best}} = \max_i y_i$, with exploration parameter $\xi \geq 0$:

\begin{equation}
I(x) = \mu(x) - f_{\text{best}} - \xi
\end{equation}
\begin{equation}
z = \frac{I(x)}{\sigma(x)}\
\end{equation}
\begin{equation}
\mathrm{EI}(x) = I(x)\Phi(z) + \sigma(x)\phi(z)
\end{equation}

where $\Phi$ and $\phi$ denote the cumulative distribution function (CDF) and probability density function (PDF) of the standard normal distribution, respectively. The candidate selected for evaluation is the graph that maximizes the Expected Improvement over the sampled candidate set:

\begin{equation}
x^{\star} = \operatorname*{arg\,max}_{x} \mathrm{EI}(x).
\end{equation}

The selected graph topology $x^{\star}$ is subsequently evaluated and added to the population for future optimization iterations \cite{frazier2018tutorial}.

\newpage
\section{Tabular Prediction Datasets}
\label{Datasets Tabular Prediction}
\begin{table}[htbp]
\centering
\caption{Overview of the IID datasets from BeyondArena~\cite{purucker2026iidgeneraltabularfoundation}, including the number of rows.}
\label{tab:iid_datasets}
\begin{tabular}{lr}
\hline
\textbf{Dataset Name} & \textbf{Number of Rows} \\
\hline
airfoil\_self\_noise & 1,503 \\
concrete\_compressive\_strength & 1,030 \\
healthcare\_insurance\_expenses & 1,338 \\
miami\_housing & 13,776 \\
superconductivity & 21,263 \\
cirrhosis\_patient\_survival\_prediction & 161 \\
body\_density\_prediction & 252 \\
blood\_tests\_drink\_prediction & 345 \\
obesity\_estimation & 498 \\
forest\_fires & 517 \\
qsar\_aquatic\_toxicity & 546 \\
student\_portuguese\_performance & 649 \\
mutual\_funds\_india & 793 \\
qsar\_fish\_toxicity & 908 \\
wine\_world\_cost & 1,279 \\
fiat\_500 & 1,538 \\
santander\_transaction\_value & 4,447 \\
qsar\_tid\_11 & 5,741 \\
wine\_quality & 6,497 \\
immoscout\_german\_house\_prices & 10,317 \\
\hline
\end{tabular}
\end{table}

\begin{table}[htbp]
\centering
\caption{Overview of the temporal datasets from BeyondArena~\cite{purucker2026iidgeneraltabularfoundation}, including the number of rows.}
\label{tab:temporal_datasets}
\begin{tabular}{lr}
\hline
\textbf{Dataset Name} & \textbf{Number of Rows} \\
\hline
garments\_worker\_productivity & 1,197 \\
coffee\_rating\_prediction & 2,369 \\
mercedes\_benz\_greener\_manufacturing & 4,204 \\
sberbank\_housing\_market\_forecasting & 27,195 \\
california\_house\_prices\_2020 & 41,528 \\
\hline
\end{tabular}
\end{table}
We use datasets from the BeyondArena ~\cite{purucker2026iidgeneraltabularfoundation} benchmark. The individual datasets remain subject to the licenses and terms of their respective original sources. We do not redistribute the underlying datasets; only aggregate benchmark results are reported.
\newpage
\section{Additional Tabular Prediction Results}
\label{Additional Results Tabular Prediction}

\begin{table}[htbp]
\centering
\caption{Results on the temporal datasets.}
\label{tab:temporal_results}
\begin{tabular}{lrrrr}
\hline
\textbf{Dataset} &
\textbf{Evolved} &
\textbf{Single Agent} &
\textbf{Plan.-Ex.-Val.} &
\textbf{Raw TabPFN} \\
\hline
california\_house\_prices\_2020
& 0.940543 & 0.940592 & 0.940009 & 0.941049 \\

coffee\_rating\_prediction
& 0.111621 & -0.252738 & 0.005208 & 0.203051 \\

garments\_worker\_productivity
& 0.444023 & 0.453376 & -0.404151 & 0.382659 \\

mercedes\_benz\_greener\_manu.
& 0.580626 & 0.580993 & 0.580140 & 0.578253 \\

sberbank\_housing\_market\_fore.
& 0.641153 & 0.640405 & 0.634187 & 0.572113 \\
\hline
\end{tabular}
\end{table}

\begin{table}[htbp]
\centering
\caption{Results on the IID datasets.}
\label{tab:iid_results}
\begin{tabular}{lrrrr}
\hline
\textbf{dataset} &
\textbf{Evolved} &
\textbf{Single Agent} &
\textbf{Plan.-Ex.-Val.} &
\textbf{Raw TabPFN} \\
\hline
airfoil\_self\_noise
& 0.975258 & 0.975962 & 0.956852 & 0.971863 \\

blood\_tests\_drink\_prediction
& 0.129454 & 0.118084 & 0.118628 & 0.118950 \\

body\_density\_prediction
& 0.692294 & 0.692378 & 0.676094 & 0.688951 \\

cirrhosis\_patient\_survival\_pred.
& 0.263664 & 0.259219 & 0.274007 & 0.318891 \\

concrete\_compressive\_strength
& 0.951657 & 0.951404 & 0.943976 & 0.951536 \\

fiat\_500
& 0.853831 & 0.855042 & 0.846765 & 0.855032 \\

forest\_fires
& 0.012283 & 0.015130 & 0.012023 & 0.007024 \\

healthcare\_insurance\_expenses
& 0.830230 & 0.830769 & 0.829713 & 0.830451 \\

immoscout\_german\_house\_pri.
& 0.697193 & 0.699006 & 0.237424 & 0.694509 \\

miami\_housing
& 0.952044 & 0.951738 & 0.951409 & 0.952178 \\

mutual\_funds\_india
& 0.777890 & 0.320690 & 0.221254 & 0.789952 \\

obesity\_estimation
& 0.109408 & 0.103002 & -0.025905 & 0.161054 \\

qsar\_aquatic\_toxicity
& 0.649739 & 0.646528 & 0.656876 & 0.660538 \\

qsar\_fish\_toxicity
& 0.621777 & 0.637216 & 0.615401 & 0.637216 \\

qsar\_tid\_11
& 0.781384 & 0.784702 & 0.778486 & 0.784864 \\

santander\_transaction\_value
& 0.425438 & 0.423872 & 0.421000 & 0.423949 \\

student\_portuguese\_per.
& 0.340675 & 0.341694 & 0.337756 & 0.340228 \\

superconductivity
& 0.931562 & 0.931529 & 0.931584 & 0.931741 \\

wine\_quality
& 0.522284 & 0.521495 & 0.508275 & 0.558394 \\

wine\_world\_cost
& 0.698428 & 0.699150 & 0.248855 & 0.714463 \\
\hline
\end{tabular}
\end{table}
\begin{table}[h]
\centering

\begin{minipage}{0.45\textwidth}
\centering
\captionof{table}{Mean $R^2$ performance of the evolved architecture and the baseline methods on the 20 IID regression datasets.}
\begin{tabular}{@{}lr@{}}
\toprule
\textbf{Model} & \textbf{Mean $R^2$} \\
\midrule
\textbf{Evolved architecture} & \textbf{0.611} \\
Single Agent (no type) & 0.588 \\
(Planner $\rightarrow$ Executor $\rightarrow$ Validator) & 0.527 \\
\bottomrule
\end{tabular}
\label{fig:IDD_mean_appendix}
\end{minipage}
\hfill
\begin{minipage}{0.48\textwidth}
\centering
\captionof{table}{Mean $R^2$ performance of the evolved architecture and the baseline methods on the 5 temporal regression datasets.}
\begin{tabular}{@{}lr@{}}
\toprule
\textbf{Model} & \textbf{Mean $R^2$} \\
\midrule
\textbf{Evolved architecture} & \textbf{0.544} \\
Single Agent (no type) & 0.473 \\
(Planner $\rightarrow$ Executor $\rightarrow$ Validator) & 0.351 \\
\bottomrule
\end{tabular}
\label{fig:temporal_mean_appendix}
\end{minipage}

\end{table}
\newpage
\section{Relative $R^2$ Difference Between Evolved and Raw TabPFN on Temporal Datasets}
\label{Relative $R^2$ Difference}
\begin{figure}[h]
\centering
\includegraphics[width=0.8\textwidth]{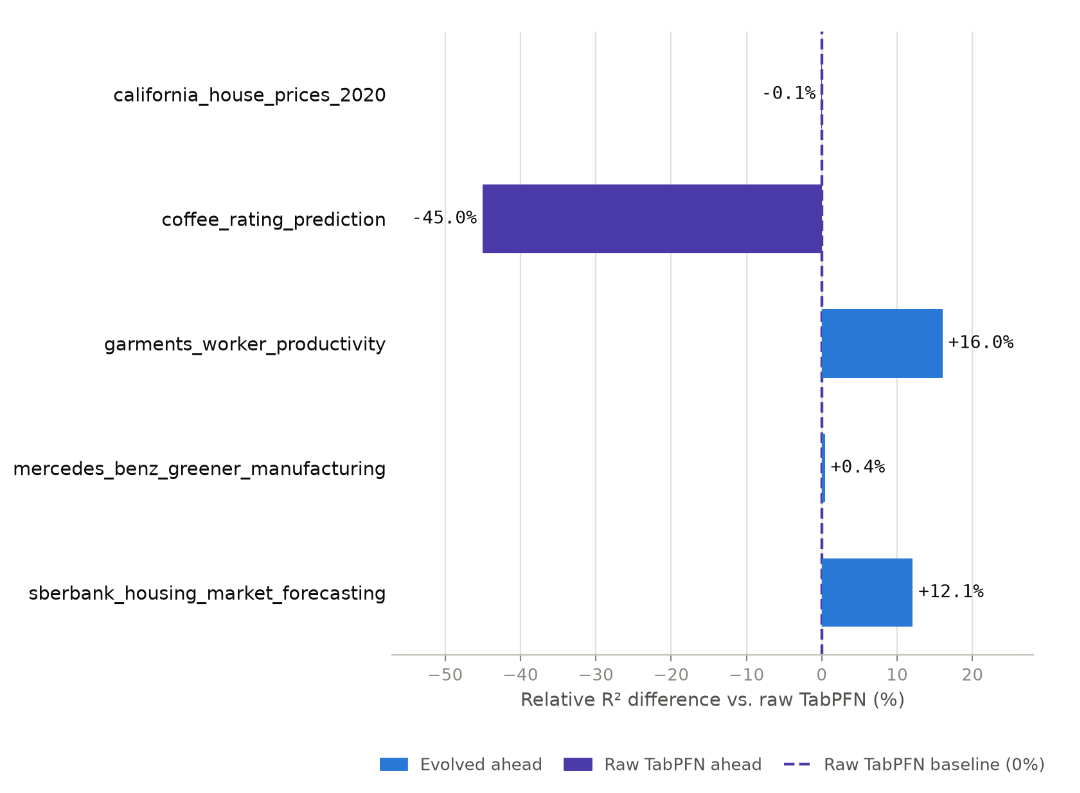}
\caption{Relative $R^2$ difference between the evolved architecture and the raw TabPFN baseline on the 5 temporal regression datasets.}
\label{fig:evolved_vs_rawtabpfn}
\end{figure}

\newpage
\section{Additional BigCodeBench Results}
\label{Additional Results BigCodeBench}
\begin{figure}[htbp]
    \centering
    \includegraphics[width=\textwidth]{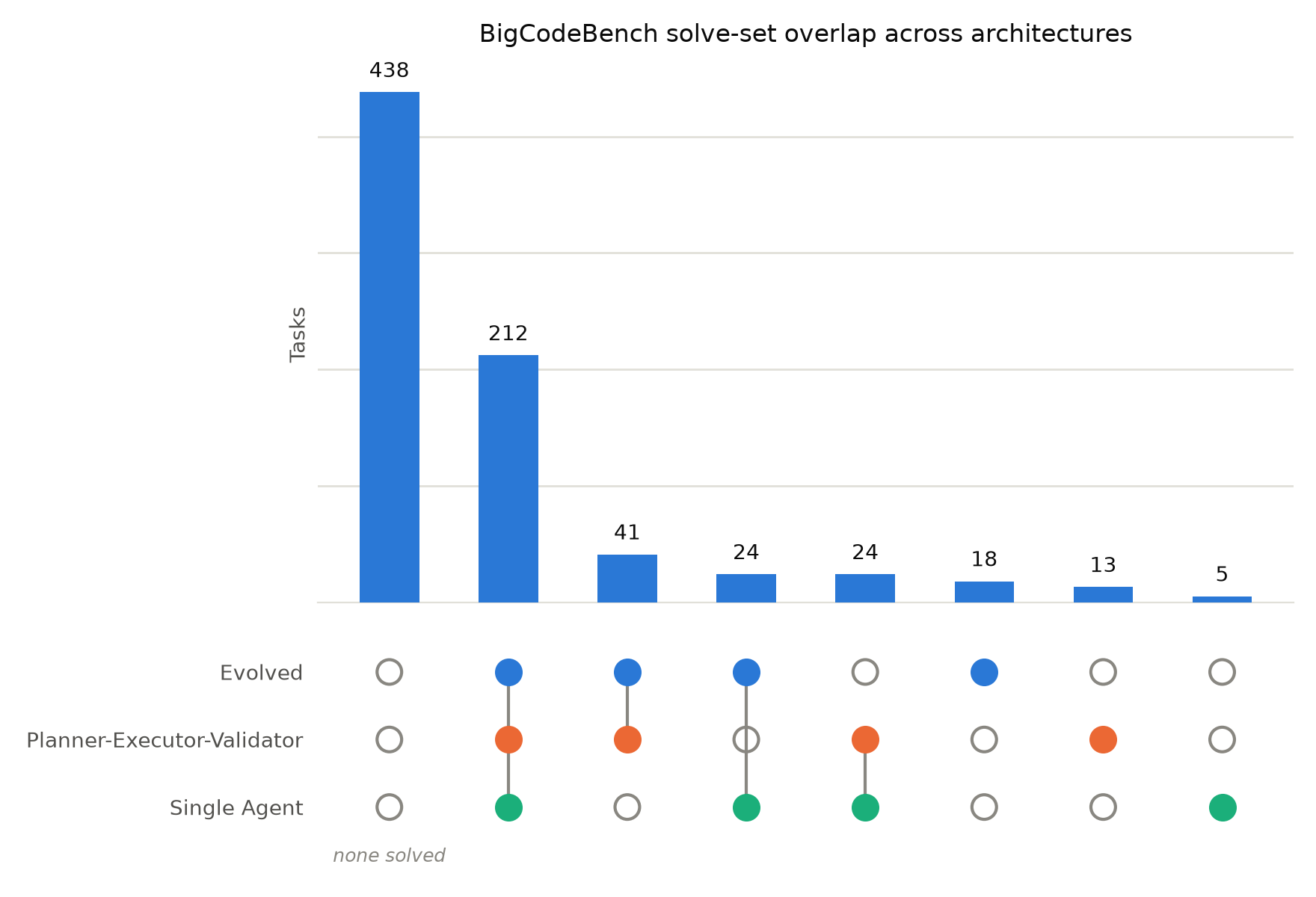}
    \caption{BigCodeBench solve-set overlap across the three considered architectures: Evolved, Planner-Executor-Validator, and Single Agent. Each bar represents the number of tasks belonging to a specific solve-set combination, while the connected markers indicate which architectures successfully solved the corresponding tasks. The largest group consists of 438 tasks that were not solved by any of the three architectures. A further 212 tasks were solved by all three architectures. Among the pairwise overlaps, 41 tasks were solved by both Evolved and Planner-Executor-Validator but not by Single Agent, while 24 tasks were solved by both Evolved and Single Agent and another 24 tasks were solved by both Planner-Executor-Validator and Single Agent. Considering tasks solved exclusively by a single architecture, Evolved solved 18 tasks, Planner-Executor-Validator solved 13 tasks, and Single Agent solved 5 tasks.}
    \label{fig:bigcodebench_solve_overlap}
\end{figure}

\newpage
\section{Architecture Search Configurations}
\label{Architecture Search Configurations}

\begin{table}[htbp]
    \centering
    \caption{Evolutionary search configuration for BigCodeBench.}
    \label{tab:evo-config-bigcodebench}
    \begin{tabular}{ll}
        \toprule
        \textbf{Parameter} & \textbf{Value} \\
        \midrule
        Dataset / task            & BigCodeBench (code generation) \\
        Population size           & 5 \\
        Generations               & 10 \\
        Crossover / mutation prob.\ & 0.5 / 0.5 \\
        Candidates per step       & 50 \\
        Search space (nodes / edges) & 2--6 / $\leq$9 (acyclic) \\
        Task LLM                  & Gemma-4-31B (temp.\ 0.25) \\
        Meta LLM                  & Gemma-4-31B (temp.\ 0.25) \\
        \bottomrule
    \end{tabular}
\end{table}
\begin{table}[htbp]
    \centering
    \caption{Evolutionary search configuration for the TabularData run.}
    \label{tab:evo-config-tabulardata}
    \begin{tabular}{ll}
        \toprule
        \textbf{Parameter} & \textbf{Value} \\
        \midrule
        Dataset / task            & TabularData (SQL feature engineering) \\
        Population size           & 5 \\
        Generations               & 10 \\
        Crossover / mutation prob.\ & 0.5 / 0.5 \\
        Candidates per step       & 50 \\
        Search space (nodes / edges) & 2--6 / $\leq$9 (acyclic) \\
        Task LLM                  & Gemma-4-31B (temp.\ 0.25) \\
        Meta LLM                  & Gemma-4-31B (temp.\ 0.25) \\
        \bottomrule
    \end{tabular}
\end{table}


\newpage

\end{document}